%% file: main.tex
\documentclass{article}

\usepackage[final]{neurips_2025_ml4ps}

\usepackage[utf8]{inputenc} 
\usepackage[T1]{fontenc}    
\usepackage{hyperref}       
\usepackage{url}            
\usepackage{booktabs}       
\usepackage{amsfonts}       
\usepackage{nicefrac}       
\usepackage{microtype}      
\usepackage{xcolor}         
\usepackage{array}
\usepackage{multirow}
\usepackage{tabularx}
\usepackage{makecell}
\usepackage[dvipsnames]{xcolor}             
\usepackage{tcolorbox}                      
\usepackage{xurl}  
\usepackage[dvipsnames]{xcolor}
\usepackage{float}

\definecolor{Blush}{HTML}{BD4257}
\definecolor{SherpaBlue}{HTML}{003D4F}

\title{The Tokenization Bottleneck: How Vocabulary Extension Improves Chemistry Representation Learning in Pretrained Language Models}

\author{Prathamesh Kalamkar\thanks{Authors contributed equally} \thanks{correspondence: prathamk@thoughtworks.com} , Ned Letcher$^{*}$, Meissane Chami$^{*}$, Sahger Lad$^{*}$, \\ {\bf Shayan Mohanty, Prasanna Pendse} \\
Thoughtworks}

\begin{document}

\maketitle

\begin{abstract}
The application of large language models (LLMs) to chemistry is frequently hampered by a "tokenization bottleneck", where tokenizers tuned on general-domain text tend to fragment chemical representations---such as SMILES---into semantically uninformative sub-tokens. This paper introduces a principled methodology to resolve this bottleneck by unifying the representation of natural language and molecular structures within a single model. Our approach involves targeted vocabulary extension—augmenting a pretrained LLM's vocabulary with chemically salient tokens, followed by continued pretraining on chemistry-domain text to integrate this new knowledge. We provide an empirical demonstration of the effectiveness of this strategy, showing that our methodology leads to superior performance on a range of downstream chemical tasks.
\end{abstract}

\input{1.introduction}
\input{2.Literature_Survey}
\input{3.Methodology}
\input{4.Results_discussion}
\input{5.Conclusion}

\begin{ack}
The authors thank NVIDIA for supporting this research.
\end{ack}

\input{6.Appendix}

\bibliographystyle{plainnat}
\bibliography{bibilography}

\end{document}

%% file: 1.introduction.tex
\section{Introduction}
Despite being powerful, LLMs typically lack the specialized domain-specific knowledge required to excel in scientific fields such as chemistry. The unique vocabulary, syntax, and concepts of chemistry necessitate a dedicated adaptation process to transform these generalist models into domain specialists. Continued pretraining (CPT) \citep{gururangan2020don} involves continuing the self-supervised pretraining objective of a base model on a new, domain-specific corpus, allowing it to acquire specialized knowledge while retaining its pre-existing capabilities. A significant challenge in continued pretraining is the phenomenon of catastrophic forgetting \citep{french1999catastrophic}, where a model's performance on general-domain tasks degrades as it continues learning from specialized-domain data.

In this paper, we introduce a principled methodology to jointly model molecular structure (SMILES) and general-purpose text by applying continued pretraining together with vocabulary extension. We also provide an empirical demonstration of its effectiveness, showing that this methodology leads to more efficient learning and superior performance on downstream tasks in the chemistry domain.

%% file: 2.Literature_Survey.tex
\section{Related Work}
The primary methods for representing molecules are string-based notations such as the Simplified Molecular Input Line Entry System (SMILES) or the more robust but less common SELFIES (SELF-referencing embedded strings). While LLMs can handle molecular text and general text \citep{sadeghi2024can}, a core challenge is that standard NLP tokenizers are ill-suited for capturing the formal grammar of these representations, leading to the semantic fragmentation that characterizes the tokenization bottleneck.

A dominant architectural paradigm addresses this challenge by treating text and molecular data as two separate modalities, enabling the use of distinct tokenizers, with approaches such as SMILES Pair Encoding handling molecular representations \citep{li2021smiles}. This approach uses a dual-encoder architecture where one encoder processes textual descriptions and another processes molecular structures. The representations from these two encoders are then aligned into a shared embedding space using techniques like contrastive learning. Prominent examples include CLAMP \citep{seidl2023enhancing}, MolFM \citep{luo2023molfm}, and PubChemSTM \citep{liu2023multi}. This approach, however, is computationally expensive and requires large, high-quality paired datasets. Another strategy is instruction tuning, where a pretrained model is fine-tuned on curated chemical-instruction datasets, such as SMolInstruct \citep{yu2024llasmol} or Mol-Instructions \citep{fang2024molinstructionslargescalebiomolecularinstruction}. This injects task-specific knowledge after pretraining, but it does not fix the underlying representational flaws learned during the initial pretraining stages. Some models, such as ChemDFM \citep{zhao2024chemdfm}, have also attempted to integrate molecular knowledge directly into a unified text-based LLM during pretraining. 

Vocabulary extension during CPT \citep{tai-etal-2020-exbert} is a domain-adaptation technique in which the vocabulary of a pretrained model is augmented with new specialized tokens before continued pretraining is performed, enabling the model to learn more semantically robust representations for domain-specific concepts. More direct, efficient, and foundational strategies for targeted vocabulary extension within existing LLM frameworks remain comparatively underexplored.

%% file: 3.Methodology.tex
\section{Methodology}

Our methodology is predicated on treating SMILES not as a distinct modality requiring a separate encoder but as a structured chemical notation that can be unified with natural language within a single representational space. This approach is motivated by the objective of creating a single cohesive model that learns a joint representation of molecular structures and natural text.

We propose a novel approach for adapting a pretrained LLM to the chemistry domain by extending its vocabulary to include chemically significant substructures before performing continued pretraining against a dataset composed of both chemistry-domain data and general-domain data. The new tokens added to the vocabulary are derived from a data-driven analysis of all SMILES strings occurring in the pretraining data, as described in Section \ref{subsection:vocabulary_extension}.

\subsection{Pretraining Data}

To enable continued pretraining, we curated a data blend from the following sources.
\begin{enumerate}
    \item \textbf{USPTO}: 
    We selected chemistry-related US patents from the USPTO dataset. Using the PatentChem library \citep{subramanian_automated_2023}, XML files were parsed into plain text, and the corresponding SMILES strings were inserted at their appropriate locations. To help the model distinguish between text and molecular modalities, all SMILES strings were enclosed within \texttt{<SMILES>} and \texttt{</SMILES>} tags.
    
    \item \textbf{Chemistry Papers}: 
    A high-quality corpus of English-language academic papers focused on chemistry was compiled from the Semantic Scholar Open Research Corpus (S2ORC) \citep{lo2020s2orc}. 
    
    \item \textbf{SMolInstruct}:
    This is a large-scale instruction-tuning dataset  for chemistry, centered on small molecules \citep{yu2024llasmol}. It features 14 distinct tasks and over three million curated samples. Given the short length of individual instruction samples, we concatenated multiple samples, separating them with \texttt{<EOS>} tokens to help the model recognize record boundaries.
    
    \item \textbf{L+M-24}: This dataset provides molecule-text pairs designed to train and evaluate a model's understanding of key natural language concepts in molecule design, such as compositionality, functionality, and abstraction \citep{edwards-etal-2024-l}.
    \item \textbf{FineWeb}: 
    To mitigate catastrophic forgetting of general knowledge---a common issue in domain adaptation---we incorporated a 10B token sample of the FineWeb dataset for data replay \citep{penedo2024fineweb}. 
\end{enumerate}

The final data blend was constructed by assigning weights to each dataset, as detailed in Table \ref{tab:data_sizes}. Higher weights were assigned to high-quality, domain-specific sources, including the chemistry papers, SMolInstruct, and L+M-24, to prioritize the learning of specialized chemical knowledge.


\newcolumntype{C}{>{\centering\arraybackslash}X}
\begin{table}[ht] 
  \caption{Pretraining data sizes and weights}
  \label{tab:data_sizes}
  \centering 
  \begin{tabularx}{\textwidth}{l *{4}{C}}
    \toprule
    \textbf{Data Source} & \textbf{Data \%} & \textbf{Data Tokens (Bn)} & \textbf{Training Data \%} & \textbf{Training Tokens (Bn)} \\
    \midrule
    FineWeb & 61\% & 10.00 & 50\% & 8.18 \\
    USPTO & 31\% & 5.00 & 35\% & 5.73 \\
    L+M-24 \& SMolInstruct & 3\% & 0.43 & 10\% & 1.64 \\
    Chemistry Papers & 6\% & 0.94 & 5\% & 0.82 \\
    \midrule 
    Total & 100\% & 16.37 & 100\% & 16.37 \\
    \bottomrule
  \end{tabularx}
\end{table}

\subsection{Vocabulary Extension}
\label{subsection:vocabulary_extension}

A core component of our methodology is the extension of the base LLM's vocabulary to incorporate important tokens for the chemistry domain. The additional tokens are derived through two distinct processes that correspond to the two modalities of text and molecular data.

\begin{enumerate}
    \item \textbf{Text modality}: The top 1,000 most frequently occurring out-of-vocabulary tokens produced by the Llama 3 tokenizer after tokenizing the blended dataset, with SMILES strings excluded.
    \item \textbf{Molecular modality}: Commonly occurring SMILES substructures extracted using the process described below, resulting in 16,795 new tokens.
\end{enumerate}



Our process for extracting new molecular tokens is based on the SMILES Pair Encoding approach, introduced by \cite{li2021smiles}, which is analogous to the Byte-Pair Encoding methodology commonly used for tokenization in LLMs \citep{sennrich-etal-2016-neural}.

\begin{enumerate}
    \item First, all SMILES strings from the USPTO, SMolInstruct, and L+M-24 datasets were extracted and filtered to remove all invalid SMILES using RDKit \citep{rdkit}.
    \item One round of data augmentation was performed by using RDKit to generate a single random alternative SMILES representation for each SMILES string. 
    \item SMILES strings were pre-tokenized into atom-level units using a regex-based tokenizer, establishing a foundation of chemically valid candidate tokens.
    \item An iterative process was then applied, merging the most frequent pairs in the token sequences. This merging continued until no candidate pair exceeded a frequency threshold of three.
\end{enumerate}

Finally, in addition to textual and molecular tokens, we added a set of special tokens required for our data format. These included tokens such as \texttt{<EOS>}, \texttt{<SMILES>}, \texttt{</SMILES>}, \texttt{<MOLFORMULA>}, and \texttt{</MOLFORMULA>}. The top ten SMILES tokens added to the vocabulary by frequency are shown in Table \ref{tab:top_tokens}. After combining these two sets of tokens, we added 17,795 new tokens to the original Llama 3 vocabulary of 128k \citep{dubey2024llama}. See Appendix \ref{app:chem_tokens} for the top ten molecular and textual tokens added.

\subsection{Continued Pretraining Setup}

To study the impact of our approach, we selected the Llama3-8B model as our base architecture. This choice was motivated by the prevalence of English in chemical literature and the demonstrated handling of catastrophic forgetting by continued pretraining of dense models using data replay, learning rate rewarming and redecaying \citep{ibrahim2024simple, parmar2024reuse, gupta2023continual}.


The addition of new tokens necessitated resizing the model's token embedding matrix. To ensure a smooth integration of the new chemical vocabulary, the embeddings for these new tokens were initialized with the mean of all existing token embeddings in the original Llama 3 model. During training, the learning rate was managed with a cosine decay schedule, following an initial warmup period, with a maximum learning rate of 3e-4. See Appendix \ref{app:hyperparams} for all hyperparameters used. 

\subsection{Experiments}
To assess the efficacy of our proposed methodology, we conducted a series of experiments centered on the continued pretraining of the Llama3-8B model. The primary comparison was between two training configurations: one with our proposed vocabulary extension and a baseline model trained without it. 
All models were trained for a single epoch on the curated pretraining data blend using a 32-GPU cluster composed of four nodes with eight NVIDIA H100s each.

\subsection{Model Evaluations}
To validate our central objective of developing a model that learns a robust joint representation of natural language and molecular structures, we evaluated our models on a suite of downstream tasks requiring a nuanced understanding of both modalities.
We selected the SMolInstruct benchmark \citep{yu2024llasmol}, a comprehensive, high-quality instruction-tuning dataset designed specifically for chemistry tasks involving small molecules, using the held-out benchmark split excluded from our pretraining. 
The tasks in this benchmark include forward synthesis (predicting reaction products), retrosynthesis (identifying reactants), molecule captioning (generating textual descriptions), name conversion (translating between molecular representations like SMILES and IUPAC), and property prediction (water solubility, octanol/water distribution coefficient, blood-brain barrier permeability,  toxicity,  HIV replication inhibition, and side effects of drugs).

We provided few-shot examples in the input prompt, which instruct the model to enclose its output in appropriate tags, such as \texttt{<SMILES>} and \texttt{</SMILES>}, to facilitate information extraction. An example prompt is shown in Appendix \ref{app:example-outputs}. These predictions were then parsed and fed into corresponding deterministic evaluation metrics for each task.

%% file: 4.Results_discussion.tex
\section{Results and Discussion}

In this section, we present the results of our evaluations as well as a brief analysis of the learned SMILES tokens that were added to the vocabulary of the customized model.   

\subsection{Evaluation Results}
\label{subsection:eval_results}

\begin{table}[!ht]
  \caption{Evaluation results for base Llama 3 and our CPT models on five task categories from the SMolInstruct dataset.}
  \label{tab:evals}
  \centering
  \small
  \setlength{\tabcolsep}{3pt}
  
  \begin{tabularx}{\textwidth}{ 
    >{\raggedright\arraybackslash}m{0.14\textwidth} | 
    >{\centering\arraybackslash}m{0.08\textwidth} | 
    >{\raggedright\arraybackslash}m{0.14\textwidth} | 
    *{6}{>{\centering\arraybackslash}X} 
  }
    \hline
        \textbf{Task} & \textbf{Samples} & \textbf{Metric} & \textbf{Llama3-8B (base)} & \textbf{Llama3-8B CPT} & \textbf{Llama3-8B CPT+ vocab ext} \\ \hline
        
        \multirow{3}{*}{\shortstack{Forward\\Synthesis}} & \multirow{3}{*}{4062} & \# Invalid & 612 & 60 & \textbf{8} \\
        ~ & ~ & \# Exact Match & 2 & 201 & \textbf{2507} \\
        ~ & ~ & Morgan FPS & 0.44 & 0.44 & \textbf{0.84}\\ \hline
        
        \multirow{3}{*}{\shortstack{Retro-\\synthesis}} & \multirow{3}{*}{4156} & \# Invalid & 1586 & 107 & \textbf{16}\\
        ~ & ~ & \# Exact Match & 0 & 22 & \textbf{1366} \\
        ~ & ~ & Morgan FPS & 0.40 & 0.31 & \textbf{0.69} \\ \hline
        
        Molecule Captioning & 2538 & METEOR & 0.19 & 0.16 & \textbf{0.22}\\ \hline
        NC-I2F & 2993 & \# Exact Match & 166 & 127 & \textbf{289} \\ \hline
        NC-I2S & 2993 & \# Exact Match & 1 & 139 & \textbf{1695} \\ \hline
        NC-S2F & 2993 & \# Exact Match & 21 & \textbf{357} & 77 \\ \hline
        NC-S2I & 2993 & \# Exact Match & 0 & 7 & \textbf{171} \\ \hline
        Property ESOL & 112 & RMSE &\textbf{1.95} & 5.42&5.68 \\ \hline
        Property LIPO & 420 & RMSE & \textbf{8.27}&59.93&31.13 \\ \hline
        Property BBBP & 197 & F1 Score & 0.11&0.79&\textbf{0.88} \\ \hline
        Property TOX & 144 & F1 Score & 0.13&0.00&\textbf{0.14} \\ \hline
        Property HIV & 4107 & F1 Score & 0.06&\textbf{0.07}&0.06 \\ \hline
        Property SIDER & 2860 & F1 Score & 0.25&0.31&\textbf{0.66}\\ \hline
  \end{tabularx}
\end{table}

The results of the different chemistry tasks used for evaluation are summarized in Table \ref{tab:evals}. These results confirm that continued pretraining (CPT) substantially improves the base model's performance on downstream chemistry tasks. Crucially, the Llama3-8B model with our proposed vocabulary extension consistently outperformed the model that underwent standard CPT, validating the benefits of integrating chemically aware tokens into the model's vocabulary.
Clear improvements over the baseline can be observed for all the tasks except the quantitative prediction tasks, such as lipophilicity (LIPO) and ESOL. 
This performance gap indicates that while the pretrained model successfully learns abstract features for SMILES understanding and classification, it struggles to learn fine-grained structural nuances required for precise numerical prediction.



\subsection{Analysis of Chemical Tokens}
\label{subsection:chem_token_analys}

Here we provide a high-level analysis of the SMILES tokens that were learned via the SMILES Pair Encoding process described in Section \ref{subsection:vocabulary_extension}.

Comparing the tokens produced by the two tokenizers over all SMILES in our corpus, the base tokenizer yields a median of 41 tokens per string, whereas the extended vocabulary yields 10. This lower fertility in the extended tokenizer is desirable, as it shortens sequences---reducing compute and fragmentation. Consistent with the goal of allocating vocabulary to frequent, reusable motifs and representing rarer forms via recomposable subunits, the extended tokenizer captures many common molecules as single tokens: 4\% of SMILES are represented by one token. See Appendix \ref{app:token_fertility} for a comparison of the full token-length distributions.

\begin{table}[!ht]
  \caption{A selection of SMILES from the dataset, comparing tokenization with the base Llama 3 tokenizer and the extended tokenizer.}
  \label{tab:chem_tokens}
  \centering
  \footnotesize
  \begin{tabular}{llll}
    \toprule
    & \textbf{SMILES string} & \textbf{Llama 3 tokenizer} & \textbf{Extended tokenizer}  \\
    \midrule
       1.
       & \texttt{N[C@@H](CCC(=O)O)C(=O)O}
       & \makecell[l]{\texttt{N}, \texttt{[C}, \texttt{@@}, \texttt{H}, \texttt{](}, \texttt{CCC}, \texttt{(=}\\ 
                      \texttt{O}, \texttt{)}, \texttt{O}, \texttt{)}, \texttt{C}, \texttt{(=}, \texttt{O}, \texttt{)}, \texttt{O}}
       & \texttt{N[C@@H](C}, \texttt{CC(=O)O)}, \texttt{C(=O)O} \\ 
       2.
       & \texttt{[1*]NC(=O)N[2*]}
       & \makecell[l]{\texttt{[}, \texttt{1}, \texttt{*}, \texttt{]}, \texttt{NC}, \texttt{(=}, \texttt{O}, \texttt{)}, \texttt{N}\\
                      \texttt{[}, \texttt{2}, \texttt{*}, \texttt{]}}
       & \texttt{[1*]N}, \texttt{C(=O)N}, \texttt{[}, \texttt{2}, \texttt{*}, \texttt{]}  \\ 
       3. 
       & \texttt{COC(=O)c1ccc(C)cc1} 
       & \makecell[l]{\texttt{C}, \texttt{OC}, \texttt{(=}, \texttt{O}, \texttt{)c}, \texttt{1}, \texttt{ccc}\\
                      \texttt{(C}, \texttt{)}, \texttt{cc}, \texttt{1}}
       & \texttt{COC(=O), c1ccc(C)cc1} \\ 
       4. 
       & \texttt{Cc1ccccc1} 
       & \texttt{Cc}, \texttt{1}, \texttt{cc}, \texttt{ccc}, \texttt{1}
       & \texttt{Cc1ccccc1} \\ 
    \bottomrule
  \end{tabular}
\end{table}

Table \ref{tab:chem_tokens} shows example SMILES from the corpus and their tokenization under both base and extended tokenizers. Beyond reduced token fertility, the extended tokenizer captures chemically meaningful substructures, often corresponding to complete or near-complete functional groups with characteristic reactivity. In the first two SMILES, we see \texttt{C(=O)O} and \texttt{C(=O)N}, corresponding to the carboxylic acid and carboxamide functional groups. The third is decomposed into \texttt{COC(=O)} and \texttt{c1ccc(C)cc1}, a methyl ester functional group and a para-methylphenyl aromatic ring. Finally, \texttt{Cc1ccccc1}, the most frequent SMILES in our corpus (the molecule toluene), is captured as a single token.

%% file: 5.Conclusion.tex
\section{Conclusion}
This work demonstrates that the tokenization bottleneck in chemistry representation learning can be mitigated through targeted vocabulary extension and continued pretraining. Our findings establish this vocabulary-centric approach as a powerful strategy for creating unified foundation models with superior performance on downstream tasks in the chemistry domain.

\section{Future Research Directions}
Integrating our continually pretrained model with external chemistry tools and reasoning frameworks, as demonstrated by systems like ChemCrow \citep{m2024augmenting} and CACTUS \citep{mcnaughton2024cactus}, presents a promising path forward. We believe that a synergistic approach that combines a semantically sound foundation model with post-training alignment and tool-use capabilities has the potential to unlock state-of-the-art performance on complex chemistry tasks.

%% file: 6.Appendix.tex
\appendix

\section{Top Chemistry Tokens}
\label{app:chem_tokens}

\begin{table}[H]
  \caption{The top 10 SMILES and textual tokens added to the new vocabulary.}
  \label{tab:top_tokens}
  \centering
  \begin{tabular}{llr}
    \toprule
    & \textbf{SMILES token}     \\
    \midrule
    1.  & \texttt{1C}           \\
    2.  & \texttt{C2}           \\
    3.  & \texttt{(=O)N}        \\
    4.  & \texttt{(=O)C}        \\
    5.  & \texttt{C)}           \\
    6.  & \texttt{C3}           \\
    7.  & \texttt{CC}           \\
    8.  & \texttt{=C1}          \\
    9.  & \texttt{c1}           \\
    10. & \texttt{ccc(}         \\
    \bottomrule
  \end{tabular}
  \begin{tabular}{llr}
    \toprule
    & \textbf{Textual token}   \\
    \midrule
    1.  & \texttt{invention}   \\
    2.  & \texttt{mmol}        \\
    3.  & \texttt{compounds}   \\
    4.  & \texttt{mixture}     \\
    5.  & \texttt{embodiments} \\
    6.  & \texttt{preferably}  \\
    7.  & \texttt{described}   \\
    8.  & \texttt{embodiment}  \\
    9.  & \texttt{herein}      \\
    10. & \texttt{thereof}     \\
    \bottomrule
  \end{tabular}
\end{table}

\section{Training Hyperparameters}
\label{app:hyperparams}

\begin{table}[H]
  \caption{Training hyperparameters for continued pretraining of Llama3-8B}
  \label{tab:hyperparams}
  \centering
  \begin{tabular}{lr}
        \toprule
        \textbf{Hyperparameters}  & \textbf{Value} \\
        \midrule
        Context Window            & 8192 \\
        Global Batch Size         & 16 \\
        Optimizer                 & Distributed Fused Adam \\
        Weight Decay              & 0.1 \\
        Betas                     & 0.9, 0.95 \\
        Scheduler                 & Cosine Annealing  \\
        Warmup Steps              & 200 \\
        Max Learning Rate         & 3e-4 \\
        Min Learning Rate         & 3e-6 \\
        \bottomrule
  \end{tabular}
\end{table}

\section{Example Outputs}
\label{app:example-outputs}

\begin{tcolorbox}[
    colback=SherpaBlue!5!white, 
    colframe=Blush,             
    title=\textbf{Example predictions for forward synthesis task}, 
    fonttitle=\bfseries\large,
    width=\textwidth,
    arc=3mm,
]
\begin{center}
    \textbf{Input Prompt}
\end{center}

\begin{verbatim}
You are a chemistry assistant. For the given reactants and reagents,
predict the resulting product. Always enclose the final SMILES string 
in <SMILES> and </SMILES> tags.

Input: <SMILES>CCOC(=N)C1=C(F)C=CC=C1F.O=C=NC1=CC(Cl)=C(OC2=NC=C(C(F)
(F)F)C=C2Cl)C(Cl)=C1</SMILES>
Based on the reactants and reagents given above, suggest a possible
product.

Answer: A possible product can be
<SMILES>CCOC(=NC(=O)NC1=CC(Cl)=C(OC2=NC=C(C(F(F)F)C=C2Cl)C(Cl)=C1)C1
=C(F)C=CC=C1F</SMILES>

Input: Propose a potential product given these reactants and reagents.
<SMILES>C1CSCCN1.CC1=CC=CC=C1.ClCCl.NC1=CNN=C1C1=NC2=CC(CN3CCOCC3)
=CC=C2N1.O=C(Cl)Cl</SMILES>

Answer:
<SMILES>O=C(NC1=CNN=C1C1=NC2=CC(CN3CCOCC3)=CC=C2N1)N1CCSCC1</SMILES>

Input:
<SMILES>COCCOCCOC1=CC=CC2=C1C(=O)N(CC1=CC=C(OC)C=C1)C2=O.[Al+3].
[H-].[Li+]</SMILES>

Answer:
<SMILES>
\end{verbatim}

\hrulefill
\par\nopagebreak
\begin{center}
    \textbf{LLM Generated Output}
\end{center}

\begin{verbatim}
COCCOCCOC1=CC=CC2=C1CN(CC1=CC=C(OC)C=C1)C2</SMILES>
\end{verbatim}
\end{tcolorbox}

\vfill

\section{Reduced Token Fertility for SMILES}
\label{app:token_fertility}

\begin{figure}[H]
  \centering
  \includegraphics[width=\linewidth]{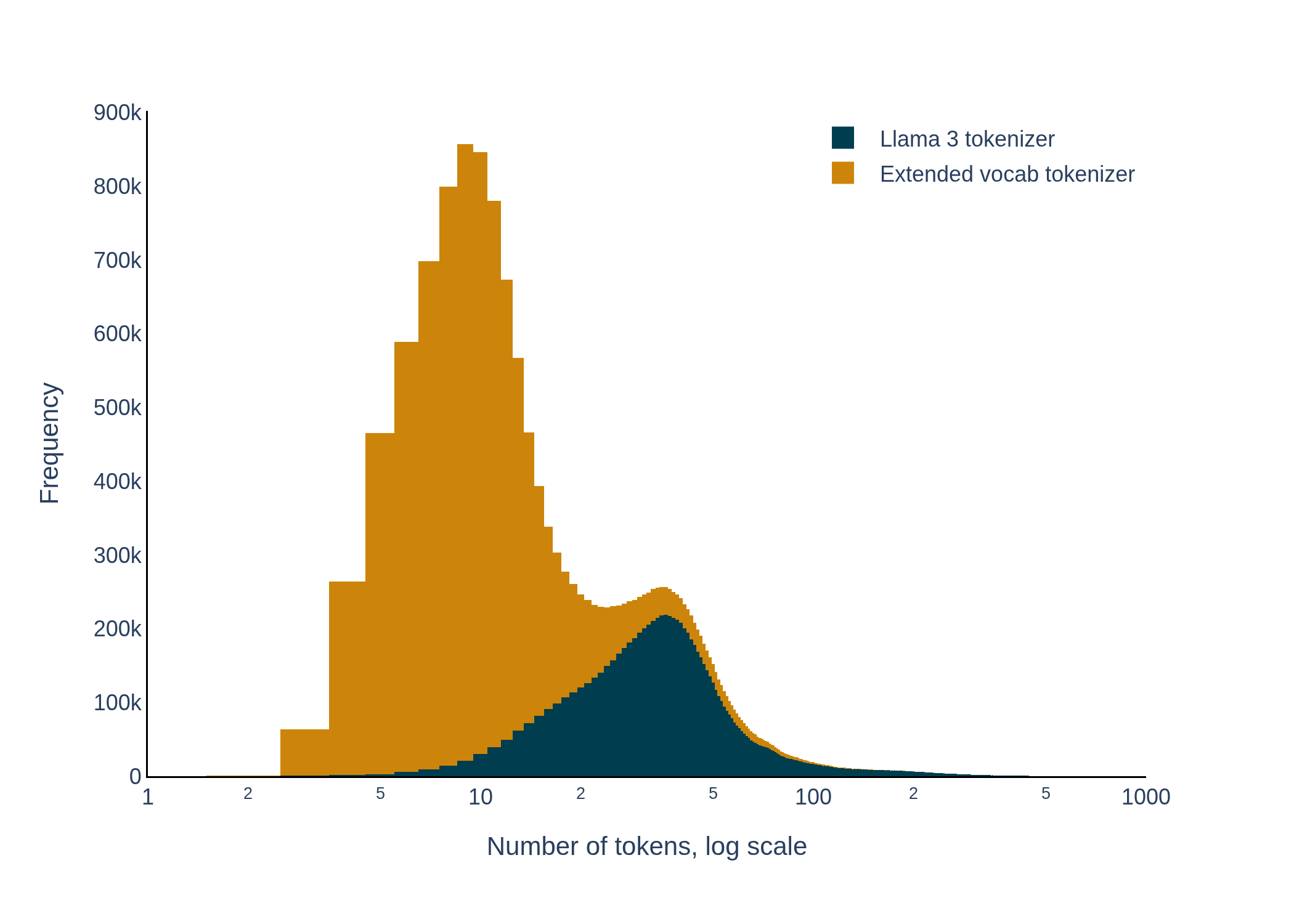}
  \caption{Histogram of the number of tokens per SMILES string}
\end{figure}